\documentclass[conference,a4paper]{IEEEtran}
\IEEEoverridecommandlockouts
\usepackage{cite}
\usepackage{amsmath,amssymb,amsfonts}
\usepackage{algorithmic}
\usepackage{graphicx}
\usepackage{multirow}
\usepackage{textcomp}
\usepackage{xcolor}
\def\BibTeX{{\rm B\kern-.05em{\sc i\kern-.025em b}\kern-.08em
    T\kern-.1667em\lower.7ex\hbox{E}\kern-.125emX}}
\begin{document}
\bibliographystyle{plain}

\title{A Compositional Feature Embedding and Similarity  Metric for Ultra-Fine-Grained Visual Categorization}


\author{\IEEEauthorblockN{Yajie Sun\thanks{This work was done during Yajie Sun's research internship at Griffith University.}, Miaohua Zhang, Xiaohan Yu, Yi Liao, and Yongsheng Gao}
\vspace{\baselineskip}

\IEEEauthorblockA{School of Engineering and Built Environment\\
    Griffith University, QLD 4111, Australia\\
    Email:\{yajie.sun; lena.zhang; xiaohan.yu; yongsheng.gao\}@griffith.edu.au; yi.liao2@griffithuni.edu.au
    }
}

\maketitle

\begin{abstract}
Fine-grained visual categorization (FGVC), which aims at classifying objects with small inter-class variances, has been significantly advanced in recent years. However, ultra-fine-grained visual categorization (ultra-FGVC), which targets at identifying subclasses with extremely similar patterns, has not received much attention. In ultra-FGVC datasets, the samples per category are always scarce as the granularity moves down, which will lead to overfitting problems. Moreover, the difference among different categories is too subtle to distinguish even for professional experts. Motivated by these issues, this paper proposes a novel compositional feature embedding and similarity metric (CECS). Specifically, in the compositional feature embedding module, we randomly select patches in the original input image, and these patches are then replaced by patches from the images of different categories or masked out. Then the replaced and masked images are used to augment the original input images, which can provide more diverse samples and thus largely alleviate overfitting problem resulted from limited training samples.  Besides, learning with diverse samples forces the model to learn not only the most discriminative features but also other informative features in remaining regions, enhancing the generalization and robustness of the model. In the compositional similarity metric module, a new similarity metric is developed to improve the classification performance by narrowing the intra-category distance and enlarging the inter-category distance. Experimental results on two ultra-FGVC datasets and one FGVC dataset with recent benchmark methods consistently demonstrate that the proposed CECS method achieves the state-of-the-art performance.
\end{abstract}


\begin{figure}[ht]
\centerline{\includegraphics[width=0.4\textwidth]{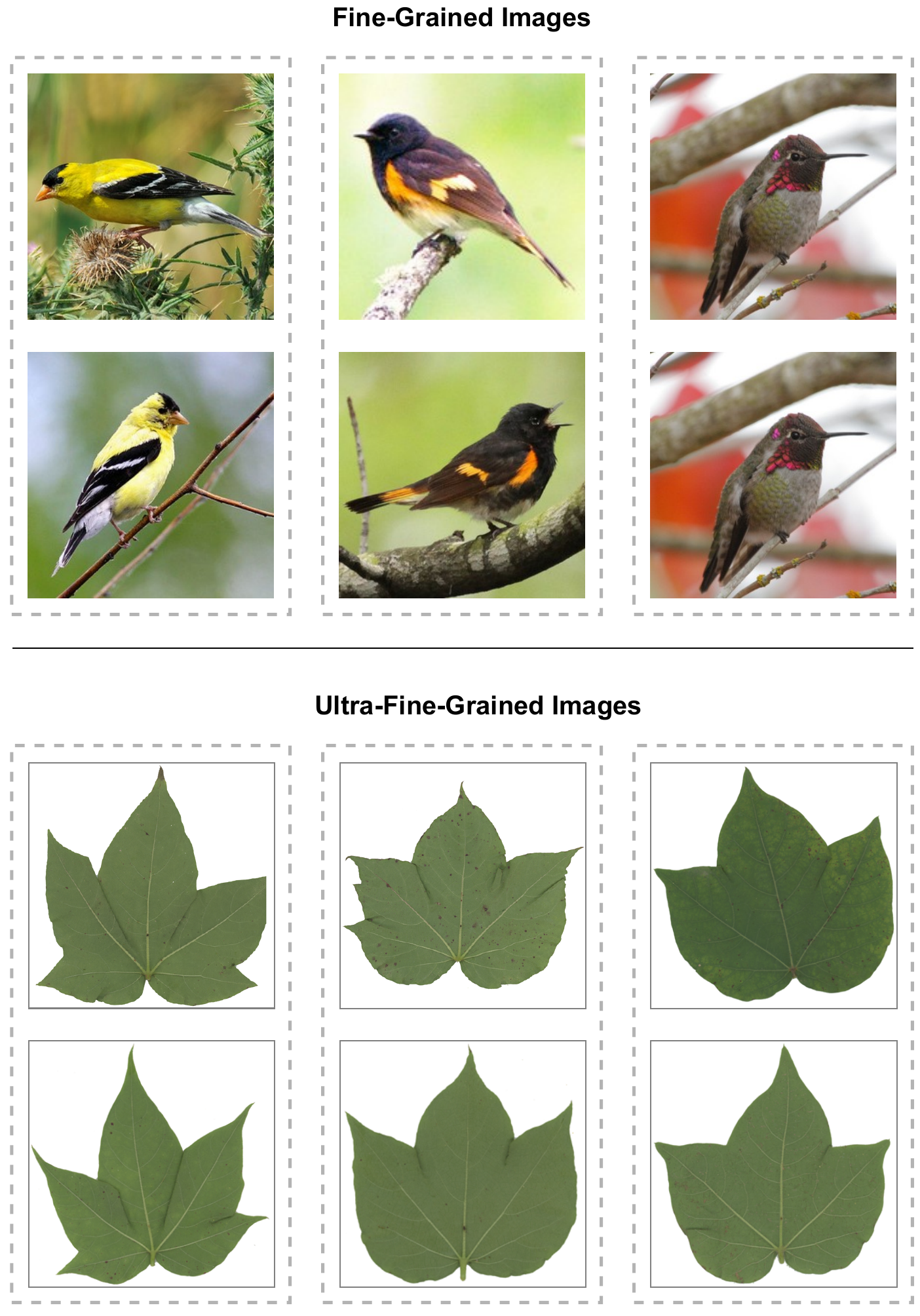}}
\caption{A data comparison of common FGVC on bird species and ultra-FGVC on cotton leaves cultivars. The birds (cotton leaves) from the same species (cultivar) in the dotted box.}
\label{dictaimage}
\end{figure}

\section{Introduction}
In computer vision, fine-grained visual categorization (FGVC) aims to classify the objects with small inter-class variances in which a clear difference may exist for different species, and has been extensively studied and made considerable progress in the past years \cite{b1},\cite{b21},\cite{b27},\cite{b28},\cite{b29},\cite{b30}. Ultra-fine-grained visual categorization (ultra-FGVC), however, focuses on classifying objects with more similar patterns among categories under a same class, and has been understudied \cite{b7},\cite{b22}. Compared  with FGVC, the inter-category differences between images in ultra-FGVC are much smaller. One representative example for ultra-FGVC tasks is to process the granularity of categorization from species level to cultivar level. A visual comparison of fine-grained images on bird species and ultra-fine-grained images on cotton leaves (cultivar) is shown in Fig.\ref{dictaimage} from which we can see that the difference of ultra-FGVC samples from different classes are too subtle to distinguish even for professional experts \cite{b25}. Thus, most fine-grained classification techniques fail to achieve ideal outcomes for ultra-FGVC tasks in practical applications, remaining a challenge problem in computer vision field. Based on current research, methods related to ultra-FGVC have advanced from manual segmentation of vein structures \cite{b5},\cite{b11},\cite{b26} to deep learning networks \cite{b7}. However, the manual method is time consuming and also highly relies on the professional knowledge of an expert, which will restrict the applications from being used by researchers with different research backgrounds. The existing deep learning based methods may suffer from image information loss due to the masking out partitioned regions directly. Thus more efforts should be made to improve the ultra-FGVC approach in aspects of both model generalization capabilities and performances.

\begin{figure}[t]
\centerline{\includegraphics[width=0.5\textwidth]{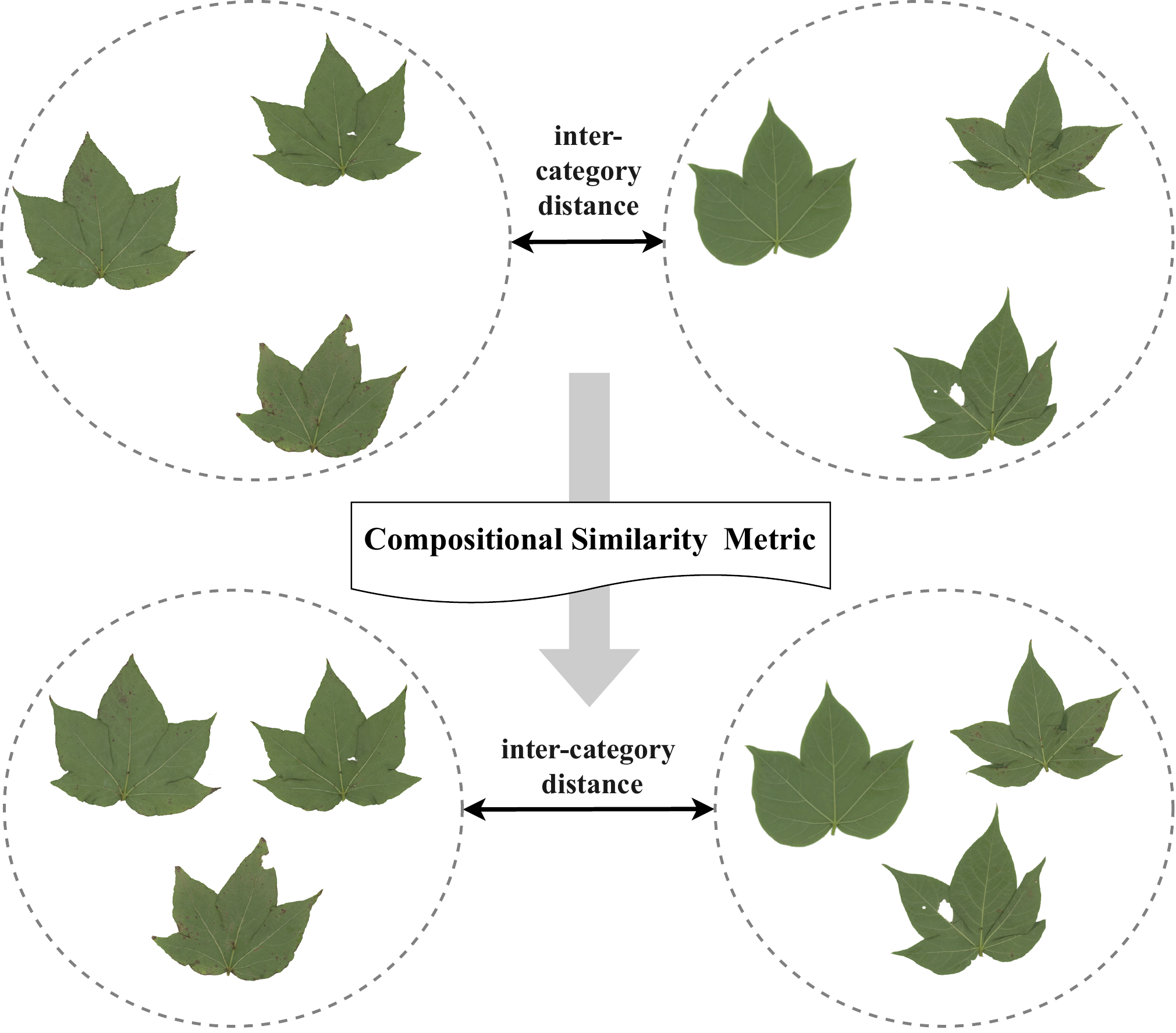}}
\caption{An example shows that how the closer intra-category distances make more obvious inter-category distances.}
\label{dictadistance}
\end{figure}

 Ultra-FGVC tasks presents two main challenges that are not addressed by existing FGVC approaches: first, the training data for each category is insufficient. Scarce sample images per category limit the scale of data available for training, which leads to a high probability of undertraining and results in severe overfitting problem. Second, the inter-class distinctions are rather tiny. In FGVC tasks, the variations of different classes are relatively obvious. For example, features used to identify different species of birds in Fig.\ref{dictaimage}, including the shape of toe, the colour of abdomen, and the texture of feather, are easy to distinguish. In contrast, distinguishing the subtle variations between categories in ultra-FGVC tasks are difficult even for human experts. To overcome the above problems, we propose a new compositional feature embedding and similarity metric (CECS) for ultra-FGVC tasks. Excepting the backbone classification network, the CECS method consists of two major components: a compositional feature embedding module and a compositional similarity metric module. To make up for the performance gap caused by insufficient training data, we first develop a compositional feature embedding module for data enhancement. In this stage, one part of the original image will be replaced with a random image form different categories or masked out. Both images preserve the ground truth label of the original image. The substitution operation generate more image samples for training, which is beneficial to preventing the occurrence of overfitting. In addition, the noisy pattern generated by the compositional feature embedding module forces the model to identify objects from the remaining partial image. The trained model will focus on more diversified and discriminative features among different categories by this way. This contributes to addressing the problem that ultra-fine-grained image features are difficult to investigate. In order to further enhance the discriminability between categories, we propose a compositional similarity metric which extracts information from the images themselves for image similarity comparison to improve performance. More specifically, we calculate the similarity of two sets of images, including the original image and the corresponding replaced image, and the replaced image and the corresponding masked image. This is conducive to recognizing similar objects within one category, encouraging the trained model pay more attention to the common features in the same category. Besides, the proposed similarity metric also improve the robustness of the model because it increases its possibility to identify the replaced image and masked one as same as the original input image. Therefore, as shown in Fig.\ref{dictadistance}, the intra-category distance can be significantly narrowed while the inter-category distance will be enlarged. It is beneficial to the subsequent classification procedure.

The contributions of this paper can be summarized as follows.
\begin{itemize}
\item A novel compositional feature embedding network is proposed to overcome the overfitting problem by augmenting the original input features. Besides, the diversity of the augmented features much improves the generalization and robustness of the model.
\item A compositional similarity metric is proposed to reduce the intra-category distance and enlarge the inter-category distance, which is beneficial to the ultra-FGVC classification performance.
\item A new loss function based on the compositional similarity metric is proposed.
\end{itemize}

\

\begin{figure*}
\centering
\includegraphics[width=0.7\textwidth]{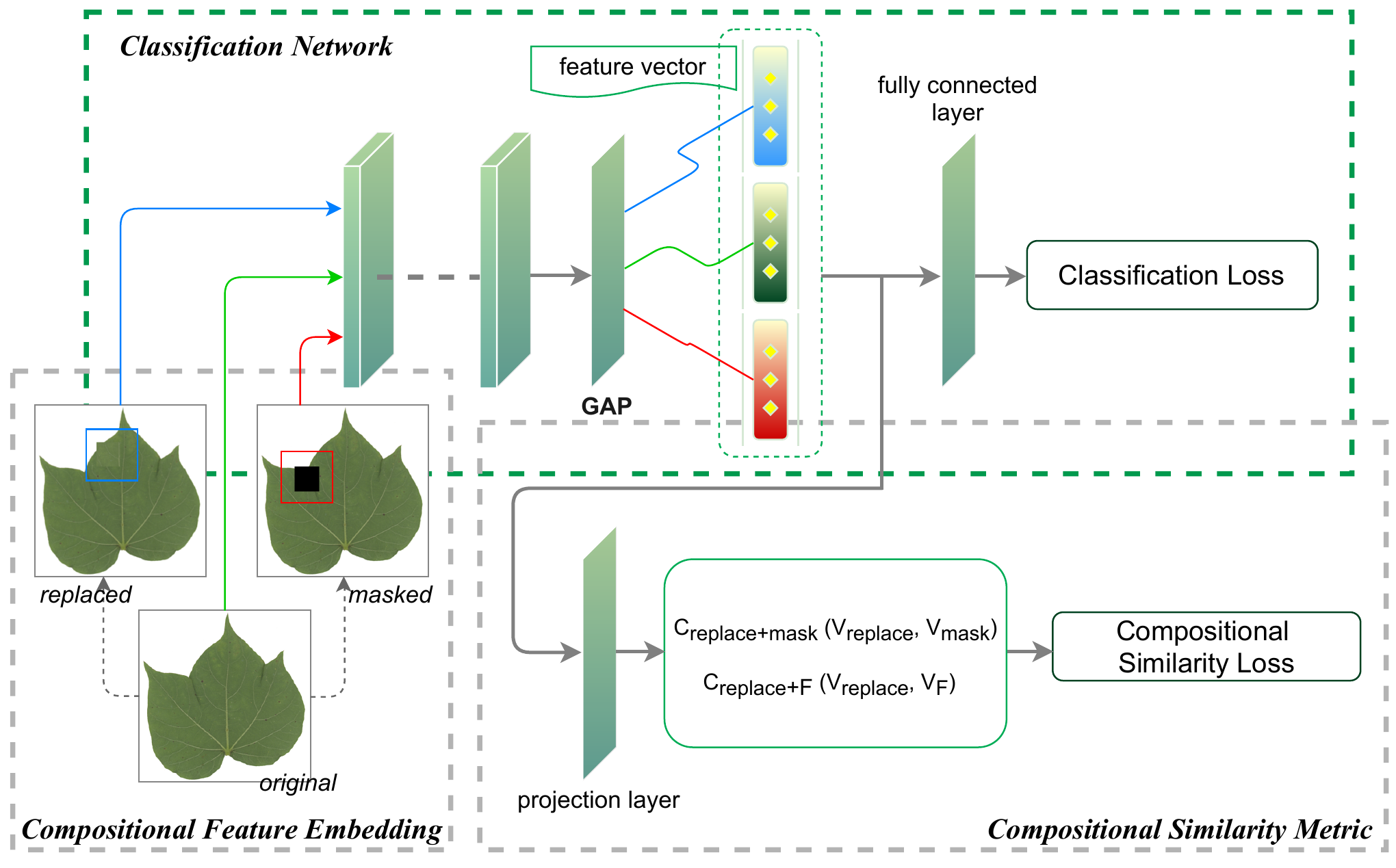}
\caption{A figure shows the whole structure of the proposed CECS method. This framework includes three parts, a compositional feature embedding module, a compositional similarity metric module, and a backbone classification network. The blue and red boxes indicate the regions to be replaced and masked, respectively.}
\label{dictaframework}
\end{figure*}
\section{Related Work}
At present, the ultra-FGVC problems have not been sufficiently studied, and still remain great challenges and at the same time provide potential opportunities for the research community. The initial research project of ultra-FGVC had been done on a new dataset containing 422 leaf images from merely three subvarieties under the same species \cite{b6},\cite{b8}. In this work, a vein features based classification method was developed by investigators to enormously improved the average classification accuracies from 55.4\% to 58.76\% compared with 41.56\% by human experts. Although their dataset is unfortunately not publicly available, their results demonstrate the possibility of classifying very fine-grained objects. Subsequently, Yu \textit{et al.} \cite{b5} developed a multiscale contour steered region integral approach. After releasing a SoyCultivarVein dataset containing 600 leaf images of 100 soybean categories, they also published another five large-scale datasets as open sources for ultra-FGVC. These datasets provided the researchers with 47,114 images of leaf samples from 3,526 different subcategories under two species of crops, cotton and soybean. Although their performance is promising, their method is not suitable for practical usage, because the method unduly dependents on manually segmented vein structures. Recently, Yu \textit{et al.}  \cite{b7} proposed a random mask covariance network (MaskCOV) without any requirement for manual pre-segmentation or extra annotations on the input images. They randomly shuffle and mask the input images, and enhance the model’s ability of learning discriminative features due to enlarging the training data volume. However, the preprocessing steps for shuffling and masking the input image in this method are cumbersome which cannot be flexibly controlled. To overcome this problem, we develop a new method that has no demand for randomly shuffling the patches of input image. In addition to masking out parts of the original image, we also replaced parts of it with the source from the random image since masking out directly may result in information loss in the original image. The method proposed in this paper also compensates for the problem of uninformative pixels during the training process caused by masked images. 

Data augmentation is generally used to overcome the overfitting problem caused by the scarcity of training samples. Cutout  \cite{b24} enlarged the scale of image data by masking out square regions of input images during training process. On this basis, CutMix  \cite{b10} cuts and pasts patches from other class of images into training images, and the ground truth labels of the two images are also mixed proportionally. Although CutMix performed well in image localization, the experimental results in image classification are not as well as expected. As an effective method for the FGVC, Decomposition and Construction Learning (DCL)  \cite{b9} approach partitions the input images into local regions of equal size. Then, a region confusion mechanism is introduced to shuffle those local regions in order to realize data augmentation at in-image level. DCL urges the model to focus on discriminative parts in the images to achieve the purpose of investigating subtle differences among different classes, and obtains remarkable performance for FGVC. In terms of the ultra-FGVC, Yu et al.  \cite{b7} recently proposed a self-supervised spatial covariance context-based (MaskCOV) method which augments images by randomly shuffling and masking partition regions of input images.

\section{Proposed Method}\label{method}
In this section, we present the proposed compositional feature embedding and similarity metric (CECS), starting with the motivation and the overview of the proposed network architecture. In addition to the backbone classification network, the CECS method consists of two key components: a compositional feature embedding module and a compositional similarity metric module. Besides a detailed introduction to the method, we also include a discussion explaining how and why the proposed method works for the ultra-FGVC tasks.

\begin{figure}
\centering
\includegraphics[width=0.5\textwidth]{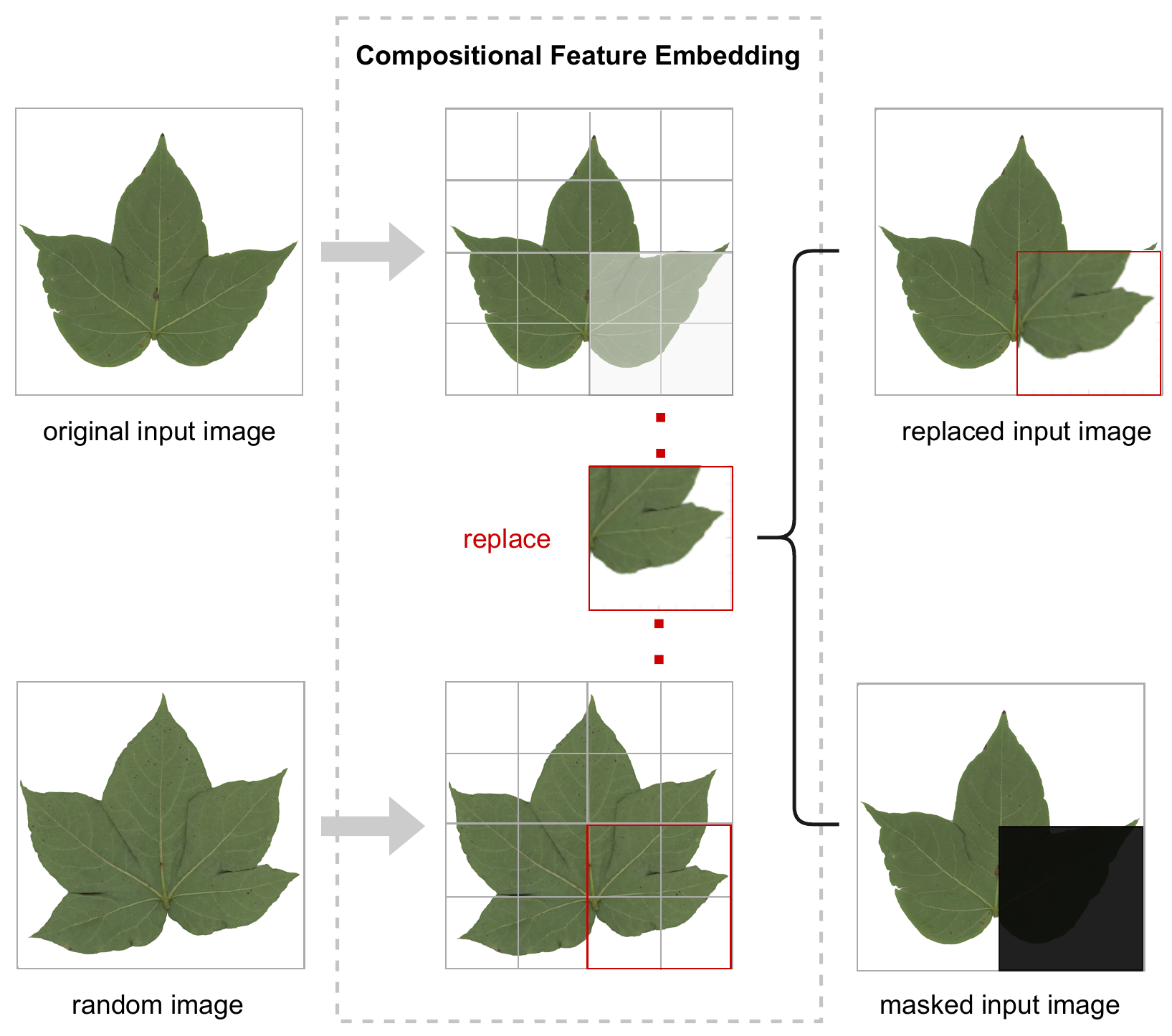}
\caption{A visual example illustrates the principle and process of the compositional feature embedding module. The region in the red box was replaced (masked).}
\label{replacemask}
\end{figure}

\subsection{Motivation}
Ultra-fine-grained visual categorization faces two tough  challenges. First of all, the recognition tasks processed by the classification network are based on scarce training samples with which the existing works are prone to severe overfitting problems \cite{b23}. To generate more training samples to alleviate overfitting, we design to replace and mask out image patches as a data augmentation approach.

Secondly, according to previous research, diversified and discriminative features are important for classifying ultra-fine-grained images. Therefore, developing a model that has the capability to focus on more discriminative features during training becomes a crucial problem. The overfitting problem indicates that the classification network may only pay attention to one or few features on the training samples during the learning process. Consequently, the model will learn and reinforce the effects of these limited features while ignoring other features. Once the images do not contain these features, the classifiers fail to make correct prediction. To address this problem, we propose the compositional feature embedding module to randomly replace and mask input images. The reason why this works for ultra-FGVC lies in that replacing or masking the images may force the model to learn more diverse features of the images, which is beneficial to improving the generalization effectiveness of the model. On this basis, a compositional similarity metric is introduced to pay more attention to common features within the same category by strengthening the intra-category similarity since closer intra-category distances would lead to more obvious inter-category distances and thus can produce better performance on ultra-FGVC tasks

\subsection{The Proposed Network Architecture}
Fig.\ref{dictaframework} illustrates the overall structure of the proposed CECS method which consists of three components, including a compositional feature embedding module, a compositional similarity metric module and a backbone classification network. During inference time, only the classification network is required for reasoning. In the following, we successively introduce the proposed compositional feature embedding module and similarity metric module.

\subsubsection{Compositional Feature Embedding}
The proposed compositional feature embedding method both enlarges the scale of data and enables the classification model to recognize discriminative and diversified features between different categories. As a data augmentation approach, it transforms the original input images into two forms: replaced input images and masked input images. Given an original input image $\textbf{\textit{F}}\in\mathbb{R}^{H\times W\times C}$, and a random image $\textbf{\textit{S}}\in\mathbb{R}^{H\times W\times C}$  from the training set. We uniformly partition the original input image $\textbf{\textit{F}}$ into \textit{n}$\times$\textit{n} equal-sized sub-regions, please see Fig.\ref{replacemask} for an example with $\textit{n}=4$. Then a  square area of size $\textit{q}\times \textit{q}$ ( $\textit{q}$ is a positive integer with $\textit{q}\in [0,  \textit{n}]$) in the original input image $\textbf{\textit{F}}$ will be replaced by a sub-image in the random image $\textbf{\textit{S}}$ with the same size and location as in $\textbf{\textit{F}}$.  The replacement $\textit{q}\times \textit{q}$ square region can be seen from the sub-image with the red box in Fig.\ref{replacemask}, and is denoted as $\textbf{\textit{I}}$ which will be further discussed in Section \ref{experimental}.

Fig.\ref{replacemask} shows a visual example demonstrating how the compositional feature embedding module performs on an original input image. For a given image $\textbf{\textit{F}}$, the first step is obtaining a random image $\textbf{\textit{S}}$ from a different category in training set. Then, a randomly selecting an area $\textbf{\textit{I}}$ to be replaced or masked in the original input image $\textbf{\textit{F}}$. For replacement stage, we extract the same area as $\textbf{\textit{F}}$ in the random image $\textbf{\textit{S}}$, and paste it into the corresponding position of the original input image $\textbf{\textit{F}}$, then the resulted new image is denoted as the replaced image. Besides replacement, a masked image by dropping out the same $\textbf{\textit{I}}$ region from original input image $\textbf{\textit{F}}$ will be obtained as well.
\

Throughout the whole process, the ground truth labels of images remain unchanged. The value on each pixel in original input image $\textbf{\textit{F}}$ and random image $\textbf{\textit{S}}$ denoted by $\textbf{\textit{F}}_{(x,y)}$ and $\textbf{\textit{S}}_{(x,y)}$ respectively with $\textit{x}\in$ [0, \textit{W}], $\textit{y}\in$ [0, \textit{H}]. The one-hot masks $\textbf{\textit{M}}_\textbf{\textit{F}}$ for $\textbf{\textit{F}}$ and $\textbf{\textit{M}}_\textbf{\textit{S}}$ for $\textbf{\textit{S}}$ can be obtained by following functions.
\begin{equation}
\text{$\textbf{\textit{M}}_\textbf{\textit{F}}$}=
\begin{cases}
\text{1},& \text{      $\textbf{\textit{F}}_{(x,y)}\nsubseteq{\textbf{\textit{I}}}$} \\
\text{0},&  \text{      $\textbf{\textit{F}}_{(x,y)}\subseteq{\textbf{\textit{I}}}$}
\end{cases}
\end{equation}
\begin{equation}
\text{$\textbf{\textit{M}}_\textbf{\textit{S}}$}=
\begin{cases}
\text{1},& \text{      $\textbf{\textit{S}}_{(x,y)}\subseteq{\textbf{\textit{I}}}$} \\
\text{0},&  \text{      $\textbf{\textit{S}}_{(x,y)}\nsubseteq{\textbf{\textit{I}}}$}
\end{cases}
\end{equation}
\

Therefore, the replaced image $\textbf{\textit{F}}_{replace}$ and masked image $\textbf{\textit{F}}_{mask}$ can be generated via the element-wise product $\odot$ between original input images and two one-hot masks as follows:
\begin{equation}
\text{$\textbf{\textit{F}}_{replace}$}=\text{ $\textbf{\textit{F}}\odot{\textbf{\textit{M}}_\textbf{\textit{F}}} + \textbf{\textit{S}}\odot{\textbf{\textit{M}}_{\textbf{\textit{S}}}}$}
\end{equation}
\begin{equation}
\text{$\textbf{\textit{F}}_{mask}$}=\text{ $\textbf{\textit{F}}\odot{\textbf{\textit{M}}_{\textbf{\textit{F}}}}$}
\end{equation}

\
In the compositional feature embedding module, all input images are augmented by randomly replacing and masking. After data scale being enlarged, each original input image in the training set will have its corresponding replaced image and masked image. The three images are then grouped together and entered into the classification model for feature extraction.
\

We adopt cross entropy as the classification loss, and the classification loss function is defined as the sum of cross entropy loss from three images: 
\begin{equation}
\text{$\textbf{\textit{L}}_{cls}$}=\text{ $\textbf{\textit{L}}_\textbf{\textit{F}} + \textbf{\textit{L}}_{replace} + \textbf{\textit{L}}_{{mask}}$}
\end{equation}

\begin{figure*}
\centering
\includegraphics[width=0.9\textwidth]{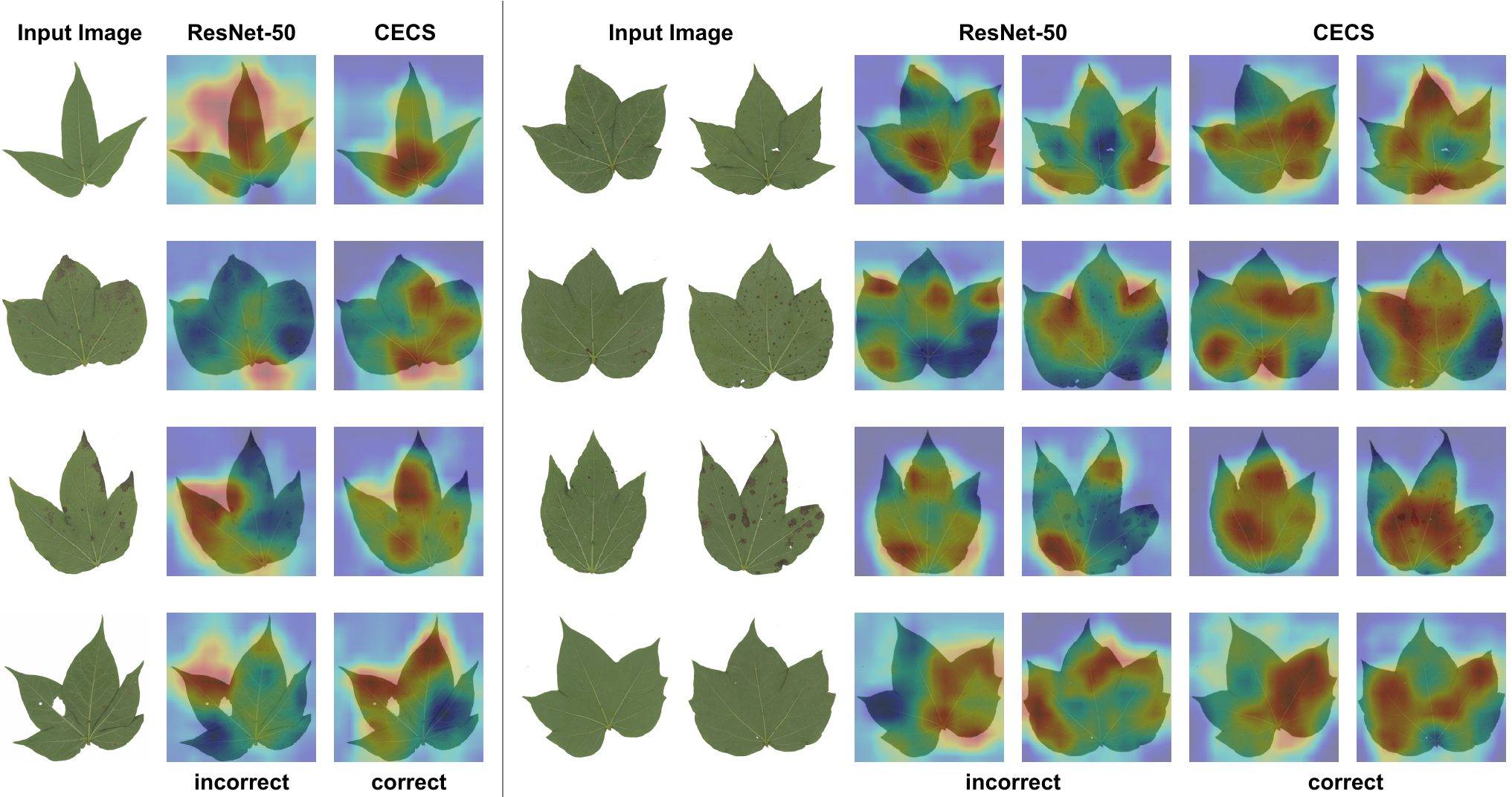}
\caption{A comparison of visualized feature maps (class activation maps) from ResNet-50 and the proposed CECS method. All predictions from the proposed CECS method are correct while from the ResNet-50 are incorrect. On the right, two cotton leaf images from each pair (per row) are  from the same category.}
\label{cam}
\end{figure*}

\subsubsection{Compositional Similarity Metric}
The random substitution scheme in the compositional feature embedding module forces feature extractor to focus on more diversified and discriminative objects in the remaining image regions, which makes the model be more effective in classifying different categories. At the same time, we aim to reduce the distance between objects under the same category, which beneficial to enlarging the distance between different categories. Thus, we develop a compositional similarity metric to investigate the similarities among various objects, and urge them more cohesiveness in the same category by continuous learning and optimization.

After feature extraction, we obtain three feature vectors, $\textbf{\textit{V}}_\textbf{\textit{F}}$, $\textbf{\textit{V}}_{replace}$ and $\textbf{\textit{V}}_{mask}$, obtained from their corresponding original input image $\textbf{\textit{F}}$, replaced image $\textbf{\textit{F}}_{replace}$ and masked image $\textbf{\textit{F}}_{mask}$. To enable the model to treat these feature vectors as the same image during the process and pay more attention to similar features, we pair these three images up for calculating the cosine similarities.
\

For replaced image $\textbf{\textit{F}}_{replace}$ and masked image $\textbf{\textit{F}}_{mask}$, the cosine similarity $\textbf{\textit{C}}_{replace+mask}$ can be calculated by,
\begin{equation}
\text{$\textbf{\textit{C}}_{replace+mask}$}=\text{ 
$\frac{{\textbf{\textit{V}}_{replace}}\cdot{\textbf{\textit{V}}_{mask}}}{\Vert{\textbf{\textit{V}}_{replace}}\Vert \cdot{\Vert{\textbf{\textit{V}}_{mask}}\Vert}}$}
\end{equation}
\

For replaced image $\textbf{\textit{F}}_{replace}$ and original image $\textbf{\textit{F}}$, the cosine similarity $\textbf{\textit{C}}_{replace+\textbf{\textit{F}}}$ is,
\begin{equation}
\text{$\textbf{\textit{C}}_{replace+\textbf{\textit{F}}}$}=\text{ 
$\frac{{\textbf{\textit{V}}_{replace}}\cdot{\textbf{\textit{V}}_\textbf{\textit{F}}}}{\Vert{\textbf{\textit{V}}_{replace}}\Vert \cdot{\Vert{\textbf{\textit{V}}_\textbf{\textit{F}}}\Vert}}$}
\end{equation}
\ 

In general, the loss function based on the compositional similarity metric is given by:
\begin{equation}
\text{$\textbf{\textit{L}}_{cos}$}=\text{ 
$(1-\textbf{\textit{C}}_{replace+mask})+(1-\textbf{\textit{C}}_{replace+\textbf{\textit{F}}})
$}
\end{equation}
\ 

With the loss obtained from the compositional feature embedding and similarity metric, the total loss in our proposed method is defined as:
\begin{equation}
\text{$\textbf{\textit{L}}$}=\text{ $\textbf{\textit{L}}_{cls} + \textbf{\textit{L}}_{cos}$}
\end{equation}

\subsection{Discussion}
In order to understand how the proposed CECS method works on the ultra-FGVC tasks, we analyse the class activation maps for ultra-fine-grained datasets. Some visual analysis from ResNet-50 and the proposed method on the samples from CottonCultivar80 are shown in Fig.\ref{cam}. We can clearly observe from the left side of Fig. 5 that many features obtained by backbone network ResNet-50 are outside of the leaves, which indicates that ResNet-50 does not learn the features of interest for the object and thus produces incorrect predictions. In contrast, the features extracted by the proposed method are mostly within the cotton leaves, which means that the proposed method is able to learn unique/discriminative features from the object rather than irrelevant information in the background. In addition, to check the effectiveness of both methods in classifying multiple samples from the same class, the feature maps for a two-sample classification task using ResNet-50 and the proposed CECS are shown on the right side of Fig. 5. The feature maps in this figure show that ResNet-50 only focus on few features while the proposed CECS method obviously learn diverse features, which indicates that the proposed method has superior ability than the baseline method in recognizing the ultra-FGVC objects of the same category.

\section{Experimental Results} \label{experimental}
To verify the effectiveness of the proposed method, in this section, we carry out extensive experiments on three publicly available datasets, including CottonCultivar80 (Cotton), SoyCultivar Local(Soy.Loc.), and Caltech-UCSD Birds (CUB)\footnote{The CUB dataset is available in http://www.vision.caltech.edu/visipedia/CUB-200.html} for ultra-FGVC tasks. The first two datasets are ultra-FGVC datasets, while the third one is usually used for FGVC tasks \cite{b3},\cite{b9}. The proposed framework is tested on different evaluation metrics and compared with recent benchmark methods.  In the following, we successively introduce the benchmark datasets, implementations details, and experimental results. Finally, we further analyse the rationality and feasibility of the proposed method via various ablation studies.
\subsection{Datasets}
$\textbf{Cotton}$: This dataset contains 480 images from 80 categories (cultivars), with 6 samples for each category. All the 80 categories belong to the same species. The significant similarities between various categories make this dataset extremely challenging, which motivates us to consider it as an ultra-fine-grained dataset. For experiments, this dataset is divided into training set and testing set with a ratio of 1:1 in the task.

$\textbf{Soy.Loc.}$: This dataset consists of  200 categories (cultivars) with 6 images per category. Hence, this dataset contains 200$\times$6=1200 images in total. All the 200 categories come from the same species. Due to very tiny difference among diverse categories, this dataset is also appropriate for ultra-FGVC investigation. For experiments, this dataset is split into training set and testing set with a ratio of 1:1 for classification.

$\textbf{CUB}$: This dataset is a public dataset which has 11,788 bird images from 200 categories. Each category contains nearly 60 samples. The CUB dataset is one of the most widely used datasets available for FGVC tasks. Different from above two datasets, the similarities between categories are more obvious than cultivars. In this dataset, 5994 images were divided into the training set, and 5794 images were divided into the testing set \cite{b3}. In this paper, we use the same data splitting protocol as in \cite{b3} for experiment implementation.

\begin{table}[htbp]
\caption{Statistics Of Datasets}
\setlength{\tabcolsep}{5mm}
\begin{center}
\begin{tabular}{|c|c|c|c|}
\hline
\textbf{Dataset} & \textbf{Category}& \textbf{Train}& \textbf{Test} \\
\hline
CottonCultivar80& 80& 240&  240\\
\hline
SoyCultivarLocal& 200& 600&  600\\
\hline
Caltech-UCSD Birds& 200& 5994&  5794\\
\hline
\end{tabular}
\label{1dataset}
\end{center}
\end{table}
\subsection{Implementation Details}
The proposed CECS method is implemented based on the Pytorch framework.  We adopt the backbone ResNet-50 as the feature extractor with the pre-trained weights obtained from the ImageNet dataset. For both the ultra-fine-grained and fine-grained classification, all input images are resized to 448$\times$448 directly. In addition, all input images are normalized and horizontally flipped with a probability of 50\% before randomly replacing and masking in the training stage. The hyper-parameter $\textit{n}$ is set to 7 for all the competing experiments. For Cotton, Soy.Loc. and CUB datasets, the hyper-parameter $\textit{q}$ related to the size of the substitution area is set to 2, 2, 1, respectively, and is then discussed in the ablation studies in Subsection $\textit{E}$ of this section.
In training stage, the proposed CECS is trained and validated for 180 epochs with a batch size of 4. We select the stochastic gradient descent (SGD) method as the optimizer with a momentum being 0.9 to optimize the learnable parameters. The initial learning rate is set to 0.0008 and then decreases by a factor of 10 every 60 epochs. All the competitive benchmarks are based on the optimal settings mentioned in the corresponding papers with discreetly fine-turning for our datasets. Besides, the classification performance is evaluated with the top-1 accuracy for all the methods in this paper.
\subsection{Evaluation on Ultra-Fine-Grained Datasets}
We first evaluate the performance of the proposed CECS approach on two ultra-fine-grained datasets for ultra-FGVC tasks, and compare it with ten benchmark methods, including Alexnet \cite{b2}, InceptionV3 \cite{b4}, Improved B-NN \cite{b14}, MobileNetV2 \cite{b12}, fast-MPN-COV \cite{b13}, VGG-16 \cite{b16}, NTS-NET7 \cite{b7}, ResNet-50 \cite{b15}, DCL \cite{b9}, and MaskCOV \cite{b7}. The classification results from different methods are shown in Table \ref{2ultra-FGVC_performance} with the best accuracy being highlighted in bold, while the second best result being underlined.
\

$\textbf{Evaluation on Cotton Dataset}$: Table \ref{2ultra-FGVC_performance} summaries the top-1 classification accuracy of 11 competing models on Cotton dataset. In addition, their respective backbones are also listed in the Table \ref{2ultra-FGVC_performance}. From the results, we can see that the proposed method achieves the best classification accuracy of 61.67\% with over 38.75\%$\sim$2.92\% higher than other approaches.

$\textbf{Evaluation on Soy.Loc. Dataset}$: All the top-1 classification accuracies of comparative methods on the Soy.Loc. dataset are listed in Table \ref{2ultra-FGVC_performance}. It can be observed that the proposed method obtains the highest classification performance of 51.33\% which it is more than 5.16\% higher than other models.
\begin{table}[t]
\setlength{\abovecaptionskip}{-2cm}
\setlength{\belowcaptionskip}{-0.2cm}
\caption{Classification Performance Of Different Methods On Cottoncultivar80 Dataset(Cotton) And Soycultivarlocal Dataset(Soy.Loc.)}
\setlength{\tabcolsep}{4.5mm}
\begin{center}
\begin{tabular}{|c|c|c|c|}
\hline
\textbf{Method}& \textbf{Backbone}& \multicolumn{2}{|c|}{\textbf{Top 1 Accuracy(\%)}} \\
\cline{3-4} 
& & \textbf{\textit{Cotton}}& \textbf{\textit{Soy.Loc.}} \\
\hline
Alexnet\cite{b2}& Alexnet& 22.92& 19.50 \\
\hline
InceptionV3 \cite{b4}& GoogleNet& 37.50& 23.00 \\
\hline
Improved B-NN \cite{b14}& VGG-16& 45.00& 33.33 \\
\hline
MobileNetV2 \cite{b12}& MobileNet& 49.58& 34.67 \\
\hline
fast-MPN-COV \cite{b13}& ResNet-50& 50.00& 38.17 \\
\hline
VGG-16 \cite{b16}& VGG-16& 50.83& 39.33 \\
\hline
NTS-NET7 \cite{b7}& ResNet-50& 51.67& 42.67 \\
\hline
ResNet-50 \cite{b15}& ResNet-50& 52.50& 38.83 \\
\hline
DCL \cite{b9}& ResNet-50& 53.75& 45.33 \\
\hline
MaskCOV \cite{b7}& ResNet-50& \underline{58.75}& \underline{46.17} \\
\hline
\textbf{Proposed Method}& ResNet-50& \textbf{61.67}& \textbf{51.33}\\
\hline
\end{tabular}
\label{2ultra-FGVC_performance}
\end{center}
\end{table}

\begin{table}[htbp]
\caption{The Classification Performance Of Different Methods On The Caltech-Ucsd Birds Dataset(CUB)}
\setlength{\tabcolsep}{7mm}
\begin{center}
\begin{tabular}{|c|c|c|}
\hline
\textbf{Method}& \textbf{Backbone}& \textbf{CUB} \\
\hline
CBP \cite{b19}& ResNet-50& 84.30  \\
\hline
RA-CNN \cite{b20}& VGG-19& 85.30 \\
\hline
ResNet-50 \cite{b15}& ResNet-50& 85.50 \\
\hline
M-CNN \cite{b18}& VGG-16-50& 85.70 \\
\hline
Improved B-NN \cite{b14}& ResNet-50& 85.80 \\
\hline
KP \cite{b17}& VGG-16& 86.20 \\
\hline
DCL \cite{b9}& ResNet-50& \textbf{86.50} \\
\hline
\textbf{Proposed Method}& ResNet-50& \underline{86.33} \\
\hline
\end{tabular}
\label{3FGVC_performance}
\end{center}
\end{table}
\subsection{Evaluation on Fine-Grained Datasets}
The experimental results shown above verify that our method has achieved state-of-the-art performance on the ultra-FGVC tasks. To further evaluate the performance of the proposed method for dealing with the general FGVC tasks, we here test it on the CUB dataset with comparisons with the published results of other seven approaches, including CBP \cite{b19}, RA-CNN \cite{b20}, ResNet-50 \cite{b15}, M-CNN \cite{b18}, Improved B-NN \cite{b14}, KP \cite{b17} and DCL \cite{b9}.

$\textbf{Evaluation on CUB Dataset}$: The top-1 classification accuracy of  all the comparative methods on this dataset are displayed in Table \ref{3FGVC_performance} from which we can see that the proposed method achieves 86.33\%, outperforming other six benchmark approaches by 2.03\%$\sim${0.13\%} and nearly matching the best performance obtained by FGVC method.

\subsection{Ablation Studies}
Having evaluated performance on different datasets with state-of-the-art benchmarks, we now present ablation studies of the proposed method on two ultra-fine-grained datasets. The proposed model consists of two modules of compositional feature embedding (CE) and compositional similarity metric (CS). Since the operation of CS is based on the enlarged data scale of module CE, we successively evaluate the performance of our method on the frameworks of backbone, backbone+CE, and backbone+CE+CS (CECS), which we denote by the module ablation studies. Besides, substitution (replace or mask) patch is also an important parameter for the proposed approach, thus the analysis for the effect of patch size on the performance of the proposed method are given in this section as well, which we denote by the substitution patch ablation studies.

$\textbf{The module ablation studies}$: This ablation studies are based on two ultra-FGVC datasets. The experimental results are listed in Table \ref{4module_cotton} and Table \ref{5module_soybean} respectively. The first part of this studies focus on verifying the effectiveness of the CE module. The results demonstrate that the CE module significantly improves the classification accuracy from 52.50\% to 60.42\% on Cotton dataset, and from 38.83\% to 50.00\% on Soy.Loc. dataset. The reason for these improvement lies  in that compared with the results from ResNet-50, the CE module provides more samples for the model in the training stage, and promotes the model to learn more discriminative and diverse features through replacing and masking. 

\begin{table}[!htb]
\caption{The Module Ablation Studies Of The Proposed Method On The CottonCultivar80 Dataset. }
\setlength{\tabcolsep}{10mm}
\begin{center}
\begin{tabular}{|c|c|}
\hline
\textbf{Method}& \textbf{Accuracy(\%)} \\
\hline
ResNet-50& 52.50  \\
\hline
ResNet-50+CE module& 60.42 \\
\hline
CECS& 61.67\\
\hline
\end{tabular}
\label{4module_cotton}
\end{center}
\end{table}
\vspace{-0.8cm}
\begin{table}[!htb]
\caption{The Module Ablation Studies Of The Proposed Method On The Soycultivarlocal Dataset. }
\setlength{\tabcolsep}{10mm}
\begin{center}
\begin{tabular}{|c|c|}
\hline
\textbf{Method}& \textbf{Accuracy(\%)} \\
\hline
ResNet-50& 38.83  \\
\hline
ResNet-50+CE module& 50.00 \\
\hline
CECS& 51.33\\
\hline
\end{tabular}
\label{5module_soybean}
\end{center}
\end{table}

The second part of module ablation studies is to verify whether the compositional feature embedding (CE) and compositional similarity metric (CS) jointly enhance the classification capability of the model. Obviously, performance is further improved by combining the two modules, which shows that learning inter-category common features has positive influence on identifying the difference between different categories.
$\textbf{The substitution patch ablation studies}$: This study focuses on the sensitivity of the model with respect to the size of the substitution patch ($\textbf{\textit{I}}$) in the CE module. As mentioned in Section \ref{method}, patch $\textbf{\textit{I}}$ is represented by $\textit{q}\times \textit{q}$ ($\textit{q}\in$ [0, \textit{n}]), and \textit{n} is set to 7. Since all input images were resized to 448$\times$448, the size of the substitution patch $\textbf{\textit{I}}$ is 64$\times$64 when \textit{q} sets to 1. In this experiment, we test the performance of the proposed method on both ultra-FGVC datasets with three different size of $\textbf{\textit{I}}$, including 64$\times$64 (\textit{q} = 1), 128$\times$128 (\textit{q} = 2), and 192$\times$192 (\textit{q} = 3).

The performance of the proposed method with different substitution size on both Cotton and Soy.Loc. datasets are listed in Table \ref{6patch_cotton} and Table \ref{7patch_soybean}, respectively. For both two datasets, the classification accuracy of the proposed method increases obviously as the size of $\textbf{\textit{I}}$ changes from 64$\times$64 to 128$\times$128. However, this trend decreases when the size of $\textbf{\textit{I}}$ is set to 192$\times$192. These results indicate that increasing or decreasing the substation size can lead to a decline in the classification accuracy. The reason for this result lies in that replacing or masking out too many patches may reduce informative regions left in the input images.

\begin{table}[!htb]
\caption{The substitution patch Ablation Studies Of The Proposed Method On The CottonCultivar80 Dataset. }
\setlength{\tabcolsep}{5mm}
\begin{center}
\begin{tabular}{|c|c|c|}
\hline
\textbf{Method}& \textbf{Backbone}& \textbf{Accuracy \%} \\
\hline
ResNet-50& ResNet-50& 52.50  \\
\hline
Proposed Method (64$\times$64)& ResNet-50& 60.83 \\
\hline
Proposed Method (128$\times$128)& ResNet-50& 61.67 \\
\hline
Proposed Method (192$\times$192)& ResNet-50& 59.17 \\
\hline
\end{tabular}
\label{6patch_cotton}
\end{center}
\end{table}

\vspace{-0.2cm}
\begin{table}[!htb]
\setlength{\abovecaptionskip}{-0.2cm}
\setlength{\belowcaptionskip}{-0.2cm}
\caption{The substitution patch Ablation Studies Of The Proposed Method On The Soycultivarlocal Dataset. }
\setlength{\tabcolsep}{5mm}
\begin{center}
\begin{tabular}{|c|c|c|}
\hline
\textbf{Method}& \textbf{Backbone}& \textbf{Accuracy \%} \\
\hline
ResNet-50& ResNet-50& 38.83 \\
\hline
Proposed Method (64$\times$64)& ResNet-50& 47.83 \\
\hline
Proposed Method (128$\times$128)& ResNet-50& 51.33 \\
\hline
Proposed Method (192$\times$192)& ResNet-50& 49.17 \\
\hline
\end{tabular}
\label{7patch_soybean}
\end{center}
\end{table}

\section{Conclusion}
This paper presents a novel compositional feature embedding and similarity metric (CECS) network to solve the drawbacks the existing ultra-fine-grained visual categorization (ultra-FGVC) method in processing extremely scarce training data and data with incredibly tiny inter-category differences. Specifically, to allivate the over-fitting problem, a compositional feature embedding module is developed to augment the original images with more diverse and discriminative features created by a random substitution scheme. Our compositional feature embedding module forces model to investigate more objects in the remaining images, which is beneficial to learning the distinct features between different categories. Besides, a compositional similarity metric is developed to enlarge the inter-class variance while reducing the intra-class variance based on the learned compositional features, which assists compositional feature embedding module to improve the performance and the generalization capability of the model. The superior performance on FGVC and ultra-FGVC datasets demonstrates the effectiveness of the proposed method in solving both FGVC and ultra-FGVC. It indicates that CECS method may be a promising solution towards addressing the challenging ultra-FGVC tasks.

\bibliography{dicta}
\end{document}